\documentclass[conference]{IEEEtran}

\makeatletter

\usepackage{blindtext}
\usepackage{eso-pic}
\IEEEoverridecommandlockouts
\usepackage{cite}
\usepackage{amsmath,amssymb,amsfonts}
\usepackage{algorithmic}
\usepackage{graphicx}
\usepackage{textcomp}
\usepackage{xcolor}
\def\BibTeX{{\rm B\kern-.05em{\sc i\kern-.025em b}\kern-.08em
    T\kern-.1667em\lower.7ex\hbox{E}\kern-.125emX}}
    
\usepackage{eso-pic}

\usepackage{algorithm}
\usepackage{algorithmic}
\usepackage{textcomp}
\usepackage{hyperref}
\usepackage{cleveref}
\usepackage{bm}
\usepackage[numbers,sort&compress]{natbib}
\usepackage{multirow}
\usepackage{booktabs}

\DeclareMathOperator*{\argmin}{arg\, min}
\DeclareMathOperator{\diag}{diag}

\newcommand{\xbm}{\vect{x}}

\usepackage{todonotes}

\def\Ncal{{\mathcal{N}}}

\def\Rbb{{\mathbb{R}}}

\newcommand{\thetabm}{\bm{\theta}}
\newcommand{\varphibm}{\bm{\varphi}}

\def\fbm{{\bm{f}}}

\def\xbm{{\bm{x}}}

\def\zbm{{\bm{z}}}

\def\Ibf{{\mathbf{I}}}

\def\Sbf{{\mathbf{S}}}
\def\Ubf{{\mathbf{U}}}
\def\Xbf{{\mathbf{X}}}

\def\Zbf{{\mathbf{Z}}}

\DeclareMathOperator{\Tr}{Tr}

\def\Dcal{{\mathcal{D}}}
\def\Ecal{{\mathcal{E}}}

\def\Ncal{{\mathcal{N}}}

\def\Abf{{\mathbf{A}}}

\def\Cbf{{\mathbf{C}}}
\def\Dbf{{\mathbf{D}}}

\def\Fbf{{\mathbf{F}}}

\def\Ibf{{\mathbf{I}}}

\def\Lbf{{\mathbf{L}}}
\def\Mbf{{\mathbf{M}}}

\def\Sbf{{\mathbf{S}}}

\def\Ubf{{\mathbf{U}}}
\def\Vbf{{\mathbf{V}}}
\def\Wbf{{\mathbf{W}}}
\def\Xbf{{\mathbf{X}}}

\def\Zbf{{\mathbf{Z}}}
\newtheorem{rem}{Remark}

\def\BibTeX{{\rm B\kern-.05em{\sc i\kern-.025em b}\kern-.08em
    T\kern-.1667em\lower.7ex\hbox{E}\kern-.125emX}}

\begin{document}
\title{\vspace*{1cm} Unsupervised Feature Selection via Robust Autoencoder and Adaptive Graph Learning
}
\author{
\IEEEauthorblockN{
1\textsuperscript{st} Feng Yu,
2\textsuperscript{nd} MD Saifur R. Mazumder,
3\textsuperscript{rd} Ying Su
}
\IEEEauthorblockA{
\textit{Department of Mathematical Sciences} \\
\textit{The University of Texas at El Paso}, Texas, USA \\
\{fyu, mmazumder, ysu2\}@utep.edu
}
\and
\IEEEauthorblockN{
4\textsuperscript{th} Oscar Contreras Velasco
}
\IEEEauthorblockA{
\textit{Department of Sociology} \\
\textit{University of California, Davis}, California, USA \\
ocontrerasvel@ucdavis.edu
}
}

\maketitle

\begin{abstract}
Effective feature selection is essential for high-dimensional data analysis and machine learning. Unsupervised feature selection (UFS) aims to simultaneously cluster data and identify the most discriminative features. Most existing UFS methods linearly project features into a pseudo-label space for clustering, but they suffer from two critical limitations: (1) an oversimplified linear mapping that fails to capture complex feature relationships, and (2) an assumption of uniform cluster distributions, ignoring outliers prevalent in real-world data. To address these issues, we propose the Robust Autoencoder-based Unsupervised Feature Selection (RAEUFS) model, which leverages a deep autoencoder to learn nonlinear feature representations while inherently improving robustness to outliers. We further develop an efficient optimization algorithm for RAEUFS. Extensive experiments demonstrate that our method outperforms state-of-the-art UFS approaches in both clean and outlier-contaminated data settings.
\end{abstract}

% \copyrightnotice{XXX-X-XXXX-XXXX-X/XX/\$XX.00 ©20XX IEEE}


\section{Introduction}
In the era of big data, high-dimensional datasets have become increasingly common across various domains, including computer vision, bioinformatics, and multimedia analysis. While such data provides rich information, its high dimensionality introduces big challenges in storage, computation, and model interpretability. To mitigate these issues, dimensionality reduction techniques play a crucial role in preprocessing by transforming or selecting the most informative features while preserving essential data characteristics. Traditional methods like Principal Component Analysis (PCA)~\cite{pearson1901liii}, Linear Discriminant Analysis (LDA)~\cite{fisher1936use}, Sufficient Dimension Reduction~\cite{huang2024robust,huang2024fr} project data into a lower-dimensional space through linear or nonlinear transformations. However, these approaches often obscure the original feature meanings, making interpretation difficult in real-world applications.  

Unlike transformation-based methods, unsupervised feature selection (UFS) directly selects a discriminative subset of features from the original data without altering its structure, thereby maintaining interpretability~\cite{yang2021multiple}. However, in unsupervised learning settings, the search for
discriminative features is done blindly, without having the class labels. Hence, UFS is considered as a much harder problem~\cite{dy2004feature}.  

Existing UFS methods can be broadly categorized into three groups: Filter methods, which evaluate features based on statistical properties (e.g., variance, Laplacian score) without involving learning algorithms~\cite{he2005laplacian,tabakhi2014unsupervised}; Wrapper methods, which employ search strategies (e.g., greedy algorithms, evolutionary computation) guided by a learning model’s performance~\cite{maldonado2009wrapper,dy2004feature}; Embedding-based methods, which integrate feature selection into an optimization framework by leveraging sparsity regularization, graph learning, or matrix factorization~\cite{yang2011nonnegative,li2012unsupervised,you2023unsupervised,huang2024framework}. 

Embedding-based approaches have gained prominence in unsupervised feature selection (UFS) due to their ability to capture feature correlations and manifold structures while maintaining computational efficiency~\cite{li2019discriminative}. Recent advances integrate adaptive graph learning, non-negative matrix factorization, and discriminative constraints to enhance robustness. However, two key challenges remain unresolved in current embedding-based methods.
First, most approaches rely on pseudo-labels in a supervised manner to approximate the true labels of the raw data, typically assuming a linear relationship between features and pseudo-labels. This simplification may fail to capture complex feature interactions. Second, existing UFS methods often overlook the presence of outliers, which are common in real-world data. Although outliers may be grouped during the training, the features derived from them can introduce misleading information—contaminating the results with irrelevant patterns while obscuring the underlying data structure.

To address these two challenges, we propose a novel unsupervised feature selection framework that overcomes existing limitations by integrating a Robust Subspace Recovery (RSR) Autoencoder (AE) into an embedding framework. Our proposed algorithm, Robust Autoencoder-Unsupervised Feature Selection (RAEUFS), leverages the AE architecture to enhance performance in traditional UFS tasks. Additionally, the RSR layer in RSRAE effectively separates outliers from benign data, ensuring that the selected features accurately represent the entire dataset. 
Experimental results on benchmark datasets demonstrate that RAEUFS outperforms state-of-the-art UFS methods for the clean datasets. Notably, in the presence of outlier contamination, our approach maintains high performance, whereas competing methods exhibit significant degradation.

The main contributions of this paper include: (1) introducing an autoencoder-based framework for feature embedding of unsupervised feature selection (UFS), which achieves state-of-the-art performance; (2) investigating, for the first time, the impact of outliers in UFS, enhancing robustness in real-world scenarios; and (3) conducting extensive experiments on both benchmark datasets with ground truth and a real-world sociology dataset without ground truth, demonstrating that our proposed method, RAEUFS, effectively selects features and provides practical guidance for real-life applications.

We organize our paper as follows: \Cref{sec:related} contains the literature review. In \Cref{sec:method}, we present RAE-RM, our robust feature selection model that combines a robust autoencoder framework with latent space clustering of local geometric structures. The details of our algorithm, RAEUFS, for solving RAE-RM are provided in \Cref{sec:opt}. \Cref{sec:experiments} presents experimental results comparing RAEUFS with other methods. Finally, conclusions are drawn in \Cref{sec:conclusion}.

\section{Related Works}\label{sec:related}
Recent years have witnessed significant progress in unsupervised feature selection (UFS) through embedding-based methods. A foundational contribution is the {Spectral Feature Selection (SPEC)} framework~\cite{zhao2007spectral}, which unifies supervised and unsupervised feature selection by measuring feature relevance using pairwise instance similarities. Building on this, {Multi-Cluster Feature Selection (MCFS)}~\cite{cai2010unsupervised} employs spectral embedding to preserve data structure by optimizing feature weights in a low-dimensional space.

Despite their effectiveness, spectral clustering-based approaches face two key limitations. First, the discrete optimization of the cluster indicator matrix is NP-hard, often yielding solutions with mixed signs and poor sparsity. Second, an overemphasis on local data structures may lead to overfitting. To address these issues, relaxation techniques are commonly adopted, where the discrete label matrix is replaced by a continuous pseudo-label matrix. This relaxed formulation preserves orthogonality by constraining the solution to the Stiefel manifold~\cite{yang2011nonnegative,li2012unsupervised,you2023unsupervised}, enabling simultaneous learning of local and global discriminative structures. Moreover, clustering performance heavily depends on the quality of the similarity matrix and suboptimal similarity learning can degrade results. Recent methods mitigate this by adaptively learning the similarity matrix, optimizing local connectivity for improved clustering~\cite{nie2014clustering,nie2016unsupervised}.

While most UFS methods rely on linear relationships, a few explore non-linear mappings. For instance,~\cite{you2023unsupervised} replaces linear spectral analysis with neural networks. However, autoencoders (AEs) remain widely adopted due to their strong representation learning capabilities. By compressing input features into a low-dimensional space and reconstructing the original data, AEs have demonstrated effectiveness in UFS~\cite{han2018autoencoder,yu2019unsupervised}.

To enhance robustness, anomaly detection can be integrated to filter outliers before feature selection. Traditional methods like Principal Component Analysis (PCA) are sensitive to outliers and often fail in corrupted data scenarios. In contrast, {Robust Subspace Recovery (RSR)} provides a more resilient framework~\cite{lerman2018overview,lerman2018fast,yu2024subspace}. The {Robust Autoencoder}~\cite{zhou2017anomaly} further improves anomaly detection by utilizing the AEs. Recent advances combine AEs with RSR layers~\cite{LaiZL20}, where normal data points are mapped close to their original positions while anomalies are pushed away. 

Despite their strengths, existing methods often employ autoencoders in a simplistic manner, neglecting robustness considerations that may compromise feature selection accuracy. To bridge this gap, we propose integrating a robust AE framework into embedded UFS, enhancing both feature selection and outlier resilience.

\section{Methodology}\label{sec:method}
In this section, we first propose our robust autoencoder regression model (RAE-RM) in \Cref{sec:RAE-RM} and introduce the adaptive graph clustering technique in \Cref{sec:graph_cluster}, then the RAE-RM based UFS approached is provided in \Cref{sec:main_model}.

\subsection{Robust AE Regression Model.}\label{sec:RAE-RM}

Let $\Xbf=[\xbm_1,\ldots,\xbm_N]^\top \in \Rbb^{N\times D}$ be the data matrix and suppose these $N$ samples are sampled from $d$ classes. Let $\mathbf{Y} = [\mathbf{y}_1, \cdots, \mathbf{y}_N]^\top \in \{0,1\}^{N \times d}$ be the cluster indicator matrix, where $\mathbf{y}_i \in \{0,1\}^{d}$ is the cluster indicator vector for $\mathbf{x}_i$. The \textit{scaled cluster indicator matrix} $\Fbf$~\cite{yang2011nonnegative} is defined as $\Fbf = \mathbf{Y}(\mathbf{Y}^T\mathbf{Y})^{-\frac{1}{2}}$. 
% \begin{align}\label{eq:scaled_cluster_indicator_matrix}
%     \Fbf = \mathbf{Y}(\mathbf{Y}^T\mathbf{Y})^{-\frac{1}{2}}.
% \end{align}
Here $\Fbf=[\fbm_1, \fbm_2, \cdots, \fbm_N]^\top\in\Rbb^{N\times d}$ and $\fbm_i$ is the scaled cluster indicator of $\xbm_i$. 
% It turns out that $\Fbf^\top\Fbf=\Ibf$. 
Thus, the linear regression model for UFS based on the scaled indicator matrix $\Fbf$ was proposed as follows~\cite{yang2011nonnegative}:
\begin{align}\label{eq:UFS_scaled_ind_mat}
    \min_{\Fbf,\Wbf} \|\Xbf^\top\Wbf-\Fbf\|_F^2, \quad
    \text{s.t. }\Fbf = \mathbf{Y}(\mathbf{Y}^T\mathbf{Y})^{-\frac{1}{2}}.
\end{align}
The elements of scaled cluster indicator matrix $\Fbf$ are constrained to discrete values, making any method relying on $\Fbf$ computationally NP-hard. An intuitive approach to address this challenge is to relax $\Fbf$ from discrete values to continuous ones under the constraint $\Fbf^\top\Fbf = \Ibf$. This relaxation preserves the orthogonality property of $\Fbf$, in which case the matrix $\Fbf$ is then referred to as the \textit{pseudo-label matrix}. To model the nonlinear relation between $\Fbf$ and the data, we incorporate a framework of robust subspace recovery autoencoder (RSR-AE) into \eqref{eq:UFS_scaled_ind_mat} and propose RSR-AE based regression model (RAE-RM) as:
\begin{align}\label{eq:UFS_pseudo_label_mat}
    \min_{\Fbf,\Wbf,\Ecal,\Abf} \|\tilde{\Zbf}-\Fbf\|_F^2, \quad
    \text{s.t. }\Fbf^\top\Fbf = \Ibf,
\end{align}
where $\tilde{\Zbf} = \Zbf\Abf = \Ecal(\Xbf^\top\Wbf)\Abf\in\Rbb^{N\times d}$ represents the output of the RSR layer, which follows the encoder. The encoder, $\Ecal:\Rbb^{p} \rightarrow \Rbb^{q}$, maps a $d$-dimensional data point to a $p$-dimensional latent code. The RSR layer is a linear transformation $\Abf \in \Rbb^{q \times d}$ that further reduces the dimension to $d$.

The idea behind this framework is to embed the indicator matrix within the latent layer, rather than in the input data space, as done in the basic UFS linear regression model \eqref{eq:UFS_scaled_ind_mat}. 
As demonstrated in \cite{LaiZL20}, the RSR layer effectively separates outliers from inliers, accomplishing the UFS task while simultaneously enhancing robustness against outliers. It is worth noting that the proposed model \eqref{eq:UFS_pseudo_label_mat} differs from existing AE-based UFS methods \cite{han2018autoencoder, yu2019unsupervised, wu2021fractal} in two key aspects: (1) these methods do not incorporate the pseudo-label matrix, which is typically beneficial for UFS tasks; and (2) they do not account for robustness in their design.

\subsection{Local geometric data structure.}\label{sec:graph_cluster}

The basic RAE-RM \eqref{eq:UFS_pseudo_label_mat} is performed in the Euclidean space and fails to capture the local geometrical structure of the data, which is crucial for discriminative analysis~\cite{gui2012discriminant}. Therefore, we cluster $\tilde{\Zbf}$ based on its local geometric structure, where $\tilde{\Zbf}$ contains the information from the original data $\Xbf$. The reasons for choosing to cluster $\tilde{\Zbf}$ instead of directly clustering $\Xbf$ are twofold: first, $\tilde{\Zbf}$ is filtered by an RSR layer that can separate out the outliers, while $\Xbf$ may contain outliers that could negatively affect the clustering performance; second, $\tilde{\Zbf}$ represents the data in a lower-dimensional space, which reduces computational complexity 
% and mitigates the curse of dimensionality 
while preserving the essential structure of the data.

To this end, we assume the pseudo label matrix $\Fbf$ preserves the cluster structure of $\tilde{\Zbf}$, i.e. the labels $\fbm_i$ and $\fbm_j$ are similar if their corresponding codes, $\tilde{\zbm}_i$ and $\tilde{\zbm}_j$, are close to each other. We denote the similarity of $\tilde{\zbm}_i$ and $\tilde{\zbm}_j$ by $s_{ij}$ and define the affinity graph $\Sbf=[s_{ij}]$. 
The matrix $\Sbf$ contains the local geometric structure for all codes and can be used to control the total scaled distances for the pseudo labels, $J(\Fbf)=\sum_{i,j}s_{ij}\|\fbm_i-\fbm_j\|^2$. A useful property of $J(\Fbf)$ is that it can be rewritten as \cite{li2012unsupervised}: 
% \vspace{-0.05in}
\begin{align*}
    \min_{\Fbf}\frac{1}{2}\sum_{i,j}s_{ij}\|\fbm_i-\fbm_j\|^2=\Tr(\Fbf^\top \Lbf_{\Sbf}\Fbf),
\end{align*}
% \vspace{-0.05in}
where $\mathbf{L}_{\Sbf} = \Dbf - \Sbf$ is the Laplacian matrix and $\Dbf=\diag(d_1,\ldots,d_N)$ is the degree matrix with $d_i=\sum_{k=1}^Ns_{ik}$. 

The function $J(\Fbf)$ depends on the affinity graph $\Sbf$, while a common choice is Gaussian kernel~\cite{shi2000normalized, shi2003multiclass, zhao2007spectral}. That is, $s_{ij} = \exp \left( -{\| \tilde{\zbm}_i - \tilde{\zbm}_j \|^2}/\sigma^2 \right), \forall \tilde{\zbm}_i \in \mathcal{N}_k(\tilde{\zbm}_j) \text{ or } \tilde{\zbm}_j \in \mathcal{N}_k(\tilde{\zbm}_i)$, where $\Ncal_k(\tilde{\zbm})$ denotes the set of $k$-nearest neighbors of $\tilde{\zbm}$. However, constructing $\Sbf$ in this manner introduces two challenges. First, selecting an appropriate value for the bandwidth $\sigma$ is critical. Second, determining the number of nearest neighbors $k$ is also non-trivial. 
To address these two challenges, we use the adaptive graph construction proposed by \cite{li2018generalized}, in which the inverse of the information entropy of $\Sbf$ is utilized. Thus, we consider the following adaptive graph component in our framework:
\vspace{-0.05in}
\begin{align}\label{eq:adaptive_graph}
&\min_{\Fbf,\Sbf}\Tr(\Fbf^\top\Lbf_{\Sbf}\Fbf)+\beta\sum_{i,j=1}^N(s_{ij}\log s_{ij}), \,\,
\text{s.t. }\Fbf^\top\Fbf = \Ibf.
\end{align}

\subsection{RAE-RM based unsupervised feature selection.}\label{sec:main_model}
The proposed RAE-RM~\eqref{eq:UFS_pseudo_label_mat} in \cref{sec:RAE-RM} utilizes the framework of the RSR-AE. We first specify the details of its settings.
% We first present the RSR-AE which is the main structure of our proposed method. 
Consider the input data $\{{\xbm}_i\}_{i=1}^{N}\subset\mathbb{R}^D$, and denote its data matrix by $\mathbf{X}=[\xbm_1,\ldots,\xbm_N]^\top\in\Rbb^{N\times D}$. Let $\Wbf\in\Rbb^{D\times p}$ be the coefficient matrix and $\xbm^s_i:=\Wbf^\top\xbm_i\in\Rbb^{p}$ represent the transformed data point with $p$ selected features. 
The encoder $\mathcal{E}$ is a neural network (NN) that maps each transformed data point to its latent code $\zbm_i = \mathcal{E}(\Wbf^\top\mathbf{x}^{(t)}) \in \mathbb{R}^q$. The RSR layer is a linear transformation $\mathbf{A} \in \mathbb{R}^{p \times d}$ that reduces the dimension to $d$ and the output of RSR is $\tilde{\zbm}_i=\Abf^\top\zbm_i\in\Rbb^d$. The decoder $\Dcal$ is a NN that maps $\tilde{\zbm}_i$ to $\tilde{\xbm}_i$ in the original ambient space $\Rbb^{D}$. The forward maps in a compact form using the corresponding data matrices is given as follows:
\begin{align*}
    & \Xbf_s =\mathbf{X}^\top\Wbf\in\Rbb^{N\times p},  \mathbf{Z} = \mathcal{E}(\mathbf{X}_s)\in\Rbb^{N\times q}, \\ & \tilde{\mathbf{Z}} = \mathbf{Z}\mathbf{A}\in\Rbb^{N\times d},  \tilde{\mathbf{X}} = \mathcal{D}(\tilde{\mathbf{Z}})\in\Rbb^{N\times D}.
\end{align*}
In RSR-AE, the following two loss functions are considered:
\begin{align}
    &\ell_{AE}^{p_1}(\Ecal,\Abf,\Dcal;\Wbf)  = \sum_{i=1}^N\left\|\xbm_i-\tilde{\xbm}_i\right\|_2^{p_1}, \label{eq:losses}
\end{align}
\vspace{-0.2in}
{\small\begin{align}
    &\ell_{RSR}^{p_2}(\Abf) \!= \!\lambda_1 \!\sum_{i=1}^N\left\|\zbm_i\!-\!\Abf\Abf^\top\zbm_i\right\|_2^{p_2} \!+\! \lambda_2 \|\Abf^\top\Abf-\Ibf_d\|^2_F.\label{eq:losses2}
\end{align}}

To enhance the robustness of the AE, we set $p_1=p_2=1$ in \eqref{eq:losses} and \eqref{eq:losses2}, adopting the least absolute deviations formulation for both reconstruction and RSR. Combining the RAE-RM of \eqref{eq:UFS_pseudo_label_mat} and the adaptive graph component \eqref{eq:adaptive_graph}, we propose the RAE-RM based unsupervised feature selection model as follows: 
% {\small
% \begin{align}\label{eq:main}
%     \min_{\substack{\thetabm,\varphibm,\Abf,\\ \Wbf, \Fbf,\Sbf}}\ell_{AE}^1&+\ell_{RSR}^1+\alpha \|\Wbf\|_{2,1} +\eta\|\tilde{\Zbf}-\Fbf\|_F^2 \nonumber\\
%     &+\gamma\Big(\Tr(\Fbf^\top\Lbf_{\Sbf}\Fbf)+\beta\sum_{i,j=1}^N(s_{ij}\log s_{ij})\Big), \\
%     &\text{s.t. }\Fbf^\top\Fbf = \Ibf, \nonumber
% \end{align}}
\begin{align}
\min_{\substack{\thetabm,\varphibm,\Abf, \\ \Wbf,\Fbf,\Sbf}}
\quad
& \ell_{AE}^1 + \ell_{RSR}^1 
+ \alpha \|\Wbf\|_{2,1}
+ \eta \|\tilde{\Zbf}-\Fbf\|_F^2  \nonumber \\[-4mm]
& + \gamma \Big(
\operatorname{Tr}(\Fbf^\top \Lbf_{\Sbf} \Fbf)
+ \beta \sum_{i,j=1}^N s_{ij} \log s_{ij}
\Big) \nonumber\\
\text{s.t.}\quad
& \Fbf^\top \Fbf = \Ibf. 
\label{eq:main}
\end{align}
where $\thetabm,\varphibm$ denote the parameters of encoder and decoder, $\ell_{AE}^1,\ell_{RSR}^1$ are given by \eqref{eq:losses} and \eqref{eq:losses2} respectively with $p_1=p_2=1$, $\mathbf{L}_{\Sbf}$ is the Laplacian matrix. Here, $\ell_{2,1}$-norm regularization term on $\Wbf$ is imposed to ensure $\Wbf$ sparse in rows.

\begin{rem}
    The dimensionality of RAE-RM is determined by the network architectures of the encoder and decoder and certain constraints exist in specific components:
    \begin{itemize}
        \item The output dimension $d$ of $\tilde{\Zbf}$ is constrained to $d\geq c+1$, where $c$ is the number of clusters. This constraint ensures: (a) \textit{Cluster separation}: the latent space becomes insufficient to distinguish all clusters when $d<c$ (b) \textit{Outlier handling}: The additional dimension ($+1$) provides a dedicated subspace for outlier identification and isolation.
        \item The encoder output dimension $q$ must satisfy $q\geq d$, ensuring the RSR layer can effectively serve as a bottleneck layer that can preserve the necessary information for cluster separation.
    \end{itemize}
\end{rem}

\section{Optimization Procedure}\label{sec:opt}

The proposed minimization problem \eqref{eq:main} involves multiple variables, for which the alternating minimization (AM) method~\cite{Bertsekas1989,yu2024hyperparameter, Wright2015} (a.k.a block coordinate minimization) is particularly well-suited. Each iteration of the AM approach consists of sequential updates, where one variable is optimized while keeping the others fixed. Our proposed algorithm solving \eqref{eq:main}, called Robust AE Unsupervised Feature Selection (RAEUFS), updates the variables in the sequence $\Ecal,\Dcal,\Abf, \Wbf, \Fbf,\Sbf$ and consists of two main components: (a) Iterative backpropagation of the two loss terms $\ell_{AE}^1 + \alpha\|\Wbf\|_{2,1}$ and $\ell_{RSR}^1$ to update the RSR autoencoder parameters ($\Ecal,\Dcal,\Abf,\Wbf$); (b) Updates for $\Sbf$ and $\Fbf$, where $\Sbf$ has an analytical solution while $\Fbf$ can be obtained through a simple computational routine. 

The detailed optimization procedures for parameters $\Ecal,\Dcal,\Abf, \Wbf$ and $\Fbf,\Sbf$ are presented in Sections~\ref{sec:update_rsr_ae} and~\ref{sec:update_f_s} respectively, with the complete RAEUFS algorithm summarized in Algorithm~\ref{alg:raeufs}.

\begin{figure}[htbp]
    \centering
    \vspace{-0.15in}
    \includegraphics[width=0.49\textwidth]{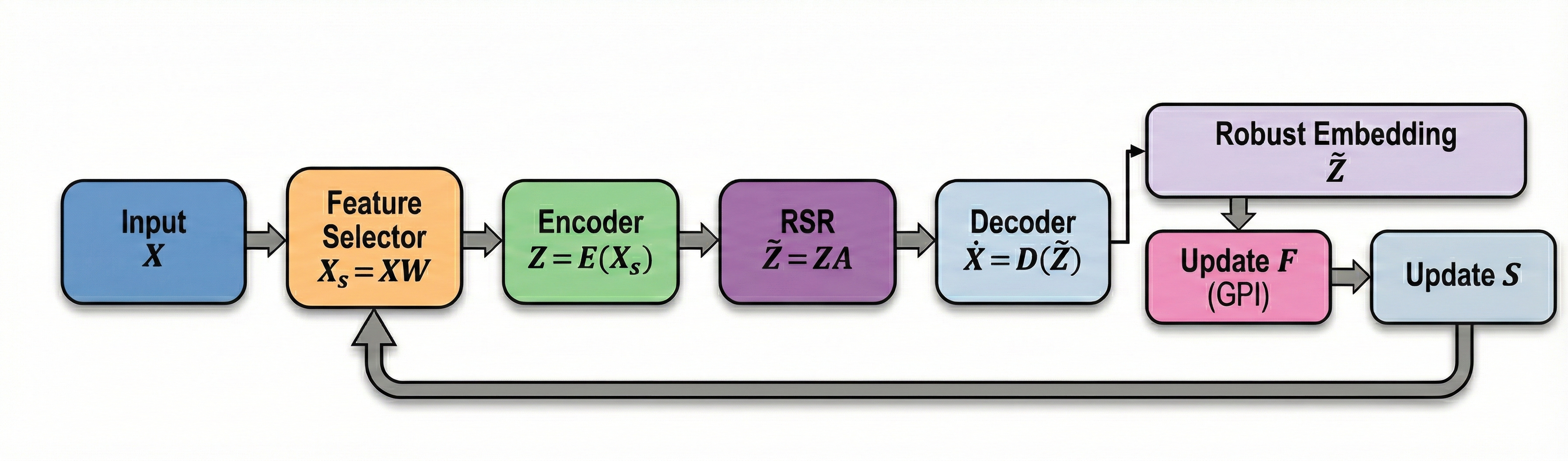}
    \caption{Overview of the RAEUFS algorithm}
    \label{fig:alg:flow}
\end{figure}
% \vspace{-0.1in}

\subsection{Updating Autoencoder.}\label{sec:update_rsr_ae}
When all other parameters ($\Abf,\Wbf,\Fbf,\Sbf$) are fixed, updating the parameters of the AE at the $k$-th iteration reduces to the following optimization problem:
\begin{align}\label{eq:update_theta_varphi}
    \min_{\thetabm,\varphibm} \ell_{AE}^{1}(\Ecal,\Abf^{(k-1)},\Dcal;\Wbf^{(k-1)})\!+\!\eta\|\tilde{\Zbf}\!-\!\Fbf^{(k\!-\!1)}\|_F,
\end{align}
which is the standard autoencoder loss function. 
We compute the gradients of the object in \eqref{eq:update_theta_varphi} with respect to $\thetabm, \varphibm$ via backpropagation. The choice of optimization method depends on the dataset size: for small datasets, we apply gradient descent (GD), while for larger datasets, we use stochastic gradient descent (SGD) or Adam~\cite{kingma2014adam}.  

Similarly, the update for $\Abf, \Wbf$ are obtained by solving the following subproblems:
{
\begin{align}
    & \argmin_{\Abf} \lambda_1 \sum_{i=1}^N\left\|\zbm_i^{(k)}\!-\!\Abf\Abf^\top\zbm_i^{(k)}\right\|_2 \!+\! \lambda_2 \|\Abf^\top\Abf\!-\!\Ibf_d\|^2_F, \label{eq:update_A}\\
    &\argmin_{\Wbf}\sum_{i=1}^N\left\|\xbm_i\!-\!\Dcal( \Abf^{(k),\top}\Ecal(\Wbf^\top \xbm_i))\right\|_2\!+\alpha \|\Wbf\|_{2,1},\label{eq:update_W}
\end{align}}
and we backpropagate the loss functions to obtain the gradients and forward with GD or Adam.
% {\small
% \begin{align}
%     \Abf^{(k)} & = \!\argmin_{\Abf} \lambda_1 \sum_{i=1}^N\left\|\zbm_i^{(k)}\!-\!\Abf\Abf^\top\zbm_i^{(k)}\right\|_2 \!+\! \lambda_2 \|\Abf^\top\Abf\!-\!\Ibf_d\|^2_F, \label{eq:update_A}\\
%     \Wbf^{(k)} & =\!\argmin_{\Wbf}\sum_{i=1}^N\left\|\xbm_i\!-\!\Dcal( \Abf^{(k),\top}\Ecal(\Wbf^\top \xbm_i))\right\|_2\!+\alpha \|\Wbf\|_{2,1},\label{eq:update_W}
% \end{align}}

% \subsection{Updating $\Fbf,\Sbf$}\label{sec:update_f_s}
\subsection{Updating \texorpdfstring{$\Fbf,\Sbf$}{F,S}.}
\label{sec:update_f_s}

The subproblem for solving $\Sbf$ have closed-form solutions, whereas that for $\Fbf$ can be addressed with a straightforward routine. Below, we first present the updates for $\Fbf$:
\begin{align}\label{QPSM}
    \min_{\Fbf^\top\Fbf=\Ibf} \eta\|\tilde{\Zbf}-\Fbf\|_F^2+\gamma\Tr(\Fbf^\top\Lbf_{\Sbf^{(k-1)}}\Fbf):=\ell_F(\Fbf).
\end{align}
Notice that the loss in \eqref{QPSM} can be rewritten as $\ell_F(\Fbf)=\Tr(\Fbf^\top(\eta\Ibf+\gamma\Lbf_{\Sbf^{(k-1)}})\Fbf-2\eta\Fbf^\top \tilde{\Zbf})+\eta\| \tilde{\Zbf}\|_F^2$, which implies that \eqref{QPSM} is a quadratic optimization problem over the Stiefel manifold (QOSM). Such problems have been extensively studied in manifold optimization literature such as \cite{journee2010generalized,absil2009optimization}.
To solve \eqref{QPSM}, we employ the Generalized Power Iteration (GPI) method~\cite{nie2017generalized}, an efficient approach for QOSM. GPI provides numerical stability and converges in few iterations. The implementation details are provided in \cref{alg:GPI}.

\begin{algorithm}[!]
\caption{GPI($\tilde{\Zbf},\Sbf$): routine for solving $\Fbf$}
\label{alg:GPI}
\begin{algorithmic}[1] 
\STATE \textbf{Input:} The matrix $\Lbf_{\Sbf}\in\Rbb^{N\times N}$, the matrix $\tilde{\Zbf}\in\Rbb^{N\times d}$ and regularization parameter $\gamma,\eta$.
\STATE \textbf{Initialize:} $\Fbf^{(0)}$ satisfying $(\Fbf^{(0)})^\top\Fbf^{(0)}=\Ibf_d$ and $\xi$ such that $\tilde{\Cbf}=\xi\Ibf_N-(\Ibf_N+\frac{\gamma}{\eta}\Lbf_{\Sbf})$ is positive definite.

\FOR{$t = 1$ to $T$}
    \STATE $\Mbf^{(t)}\leftarrow2\tilde{\Cbf}\Fbf^{(t-1)}+2
    \tilde{\Zbf}$
    \STATE  $[\Ubf,\Sbf,\Vbf]\leftarrow\text{RandomizedPCA}(\Mbf^{(t)})$~\cite{rokhlin2010randomized}
    \STATE Update $\Fbf^{(t)}\leftarrow \Ubf_{d}\Vbf^\top$ where $\Ubf_{d}$ is the matrix consisting of the first $d$ columns of $\Ubf$.
\ENDFOR
\STATE \textbf{Output:} The pseudo label matrix $\Fbf^{(T)}$.

\end{algorithmic}
\end{algorithm}

% If $\Lbf_{\Sbf}$ is the Laplacian matrix, then the following identity holds:
% \begin{align*}
%     \Tr(\Fbf^\top\Lbf_{\Sbf}\Fbf) = \frac{1}{2}\sum_{i,j=1}^N\|\fbm_{i}-\fbm_j\|_2^2s_{ij}.
% \end{align*}
When all parameters except $\Sbf$ are fixed, minimizing over $\Sbf$ in \eqref{eq:main} reduces to
\vspace{-0.1in}
\begin{align}\label{eq:minimize_S}
    \Sbf^{(k)}&=\argmin_{\Sbf\in\Rbb^{N\times N}}\sum_{i,j=1}^N(\|\fbm_{i}-\fbm_j\|_2^2s_{ij}+2\beta s_{ij}\log s_{ij}), \nonumber\\ &\text{s.t. } \sum_{j=1}^Ns_{ij}=1,\quad s_{ij}\geq 0,
\end{align}
which is true beacuse of the identity, $\Tr(\Fbf^\top\Lbf_{\Sbf}\Fbf) = \frac{1}{2}\sum_{i,j=1}^N\|\fbm_{i}-\fbm_j\|_2^2s_{ij}$.
The Lagrangian function of \eqref{eq:minimize_S} is then given by 
\begin{align*}
    \ell_S(\Sbf;\Phi,\Pi)&=\sum_{i,j=1}^N(\|\fbm_{i}-\fbm_j\|_2^2s_{ij}+2\beta s_{ij}\log s_{ij}) \\
    & +\sum_{i=1}^N\phi_i(\sum_{ij=1}^Ns_{ij}-1)-\sum_{ij=1}^N\pi_{ij}s_{ij},
\end{align*}
where $\Phi=\{\phi_i|i\in[N]\}$ and $\Pi=\{\pi_{ij}|i,j\in[N]\}$ are the Lagrangian multipliers. The Karush–Kuhn–Tucker (KKT) conditions of $\ell_S(\Sbf;\Phi,\Pi)$ yield the following equations
{\small
\begin{align*}
    \begin{cases}
        \|\fbm_{i}-\fbm_j\|_2^2+2\beta(1+\log s_{ij})+\phi_i-\pi_{ij}=0, &\forall i,j\in[N] \\
    s_{ij}\geq0,\quad \pi_{ij}\geq0, \pi_{ij}s_{ij}=0, &\forall i,j\in[N] \\
    \sum_{ij=1}^Ns_{ij}=1, &\forall i\in[N]
    \end{cases}
\end{align*}}
Its solution is given by that for $\forall i,j\in[N]$,
% {\small
% \begin{align}\label{eq:update_S}
%     s_{ij}=\exp\left(-\frac{\|\fbm_{i}-\fbm_j\|_2^2}{2\beta} \right)\Bigg/\left\{\sum_{j=1}^N\exp\left(-\frac{\|\fbm_{i}-\fbm_j\|_2^2}{2\beta}\right)\right\},
% \end{align}}
\begin{align}\label{eq:update_S}
s_{ij} = 
\exp\Big(-\frac{\|\fbm_i - \fbm_j\|_2^2}{2\beta}\Big) 
\Big/ 
\sum_{k=1}^{N} \exp\Big(-\frac{\|\fbm_i - \fbm_k\|_2^2}{2\beta}\Big).
\end{align}
\vspace{-0.1in}

\begin{algorithm}[h]
\caption{RAEUFS}
\label{alg:raeufs}
\begin{algorithmic}[1] 
\STATE \textbf{Input:} Data matrix: $\mathbf{X} \in \mathbb{R}^{N \times D}$, architecture of $\Ecal:\Rbb^p\rightarrow\Rbb^q$ and $\Dcal:\Rbb^d\rightarrow\Rbb^D$ with $d=c+1$, number of selected features: $p$, number of clusters: $c$, parameters: $\lambda_{1}, \lambda_{2}, \alpha, \beta$.

\STATE \textbf{Initialization:} Random matrices $\mathbf{W}^{(0)} \in \mathbb{R}^{D \times p}, \mathbf{A} \in \mathbb{R}^{q \times d}$, initial parameters of $\Ecal,\Dcal$.

\FOR{$k = 1$ to $K$}
    \STATE Backpropagate \eqref{eq:update_theta_varphi} w.r.t. $\thetabm,\varphibm$. Update $\thetabm^{(k)},\varphibm^{(k)}$.
    % with Adam~\cite{kingma2014adam}
    \STATE Backpropagate \eqref{eq:update_A} w.r.t. $\Abf$. Update $\Abf^{(k)}$.
    \STATE Backpropagate \eqref{eq:update_W} w.r.t. $\Wbf$. Update $\Wbf^{(k)}$.
    \STATE Update $\Fbf^{(k)}=\text{GPI}(\tilde{\Zbf}^{(k)},\Sbf^{(k-1)})$ where $\tilde{\Zbf}^{(k)}=\Ecal(\Xbf^\top\Wbf^{(k)})\Abf^{(k)}$ and GPI is provided by \cref{alg:GPI}
    \STATE Update $\Sbf^{(k)}$ by \eqref{eq:update_S}
\ENDFOR

\STATE \textbf{Output:} Weight matrix, $\mathbf{W}^{(K)} \in \mathbb{R}^{D \times p}$.
\end{algorithmic}
\end{algorithm}

% \begin{figure}[htbp]
%     \centering
%     \includegraphics[width=0.49\textwidth]{fig/alg_optimization.png}
%     \caption{Algorithm optimization details}
%     \label{fig:alg:optimization}
% \end{figure}

\section{Experiments}\label{sec:experiments}

In this section, we conduct extensive experiments to evaluate the performance of the proposed RAEUFS for feature selection in clustering tasks. The experiments consist of two main parts: datasets with ground truth labels and one real world application dataset without ground truth information.

\subsection{Datasets with Ground Truth}
 \noindent\textbf{Datasets.} We evaluate our method on six publicly available datasets, 
 % \texttt{USPS}~\cite{hull1994database}, \texttt{Jaffe}~\cite{lyons1998jaffe}, 
 % \texttt{COIL20}~\cite{nene1996coil}, 
 % \texttt{lung}~\cite{bhattacharjee2001classification}, 
 % \texttt{Isolet}~\cite{cole1991isolet}, and 
 % \texttt{WarpPIE10P}~\cite{wei2021unsupervised}. 
 their descriptions are provided in \Cref{tab:dataset_description}.
% of the Supplementary Material. All input data $\xbm$ are normalized to the range $[0,1]$.
Moreover, to assess the robustness of our algorithm, we synthetically generate outliers for each dataset by sampling from $\mathcal{N}(0, \Ibf_D)$ and incorporate them into the training process.

\begin{table}[h]
\vspace{-0.1in}
\caption{Description of the datasets.}
\label{tab:dataset_description}
\centering
\begin{tabular}{lccc}
\toprule
\textbf{Dataset}    & \textbf{Observations} & \textbf{Features} & \textbf{Clusters} \\
\midrule
lung \cite{bhattacharjee2001classification}         & 203    & 3,312 & 5   \\
Jaffe \cite{lyons1998jaffe}        & 213    & 676   & 10  \\
Isolet \cite{cole1991isolet}      & 1,560  & 617   & 26  \\
COIL20 \cite{nene1996coil}      & 1,440  & 1,024 & 20  \\
WarpPIE10P \cite{wei2021unsupervised}   & 210    & 2,420 & 10  \\
USPS \cite{hull1994database}        & 9,298  & 256   & 10  \\
\bottomrule
\end{tabular}
\end{table}

\noindent\textbf{Compared Methods.} 
We evaluate RAEUFS against two state-of-the-art UFS methods: Generalized Uncorrelated Regression with Adaptive Graph (URAFS)~\cite{li2018generalized} and Neural Networks with Self-Expression (NNSE)~\cite{you2023unsupervised}. 
% All methods will employed and then \textit{k}-means method~\cite{lloyd1982least} will be applied to the reduced feature data $\mathbf{X}_s$ with repeating 100 times. 
After employing the methods, \textit{k}-means ~\cite{lloyd1982least} will be applied to the reduced feature data $\mathbf{X}_s$ with repeating 100 times. Additionally, we include a baseline where \textit{k}-means is applied directly to the original data without any feature selection.

\noindent\textbf{Parameter Settings.} To determine the optimal parameters ($\alpha$, $\beta$, $\gamma$, $\eta$, $\lambda_1$, $\lambda_2$), we perform a grid search over the values $\{10^{-2}, 10^{-1}, 1, 10\}$ and best results are recorded. 
% The number of selected features is fixed at 200. 
% For detailed results across different parameter settings and feature counts (see \Cref{app:parameter_sensitivity}).
 
% in the Supplementary Materials. 

\noindent\textbf{Evaluation Metrics.} To comprehensively evaluate performance, we use two complementary metrics: clustering accuracy (ACC) and normalized mutual information (NMI)~\cite{fan1949theorem}. Both metrics assess clustering quality, with higher values indicating better performance. Since ACC and NMI capture different aspects of clustering effectiveness, their combined use enables a more thorough evaluation of the clustering performance. For datasets containing corrupted examples, we restrict our evaluation to only the clean data samples.

\noindent\textbf{Results.} The comparisons of RAEUFS, URAFS, NNSE as well as the vanilla \textit{k}-means for the clean and outlier-contaminated datasets are reported in \Cref{tab:benign_result} and \Cref{tab:outlier_result}. The results in \Cref{tab:benign_result} indicate that the proposed methods RAEUFS outperforms URAFS and NNSE. The two exceptions are \texttt{COIL20} in terms of ACC and \texttt{USPS} in terms of NMI, in which the accuracy of RAEUFS is very close to NNSE. 

\Cref{tab:outlier_result} shows that even the datasets are contaminated with 30\% outliers during the training, RAEUFS is able to identify the outliers and achieves similar clustering performance compared to the clean datasets, while other algorithms degenerate the performance. The only exception is \texttt{WarpPIE10P} dataset that NNSE has the highest accuracy but its accuracy is much lower than RAEUFS in the clean datasets.

\begin{table}[!ht]
\centering

\caption{Comparison of RAEUFS, URAFS, and NNSE with 200 selected features for \textbf{clean dataset}. Mean and standard deviations (in parentheses) of ACC and NMI are reported over 100 repetitions. The best performance across the algorithms for each dataset is bold underlined.
}
\label{tab:benign_result}

\resizebox{\linewidth}{!}{%
\begin{tabular}{lcccc}
\toprule
\multirow{2}{*}{Dataset} 
  & \multicolumn{4}{c}{Accuracy (ACC) (\%)} \\
\cmidrule(lr){2-5}
  & Baseline & RAEUFS & URAFS & NNSE \\
\midrule
\texttt{lung}       & 66.50 (0.10) & \underline{\textbf{71.08 (0.02)}} & 64.88 (2.68) & 68.72 (0.36) \\
\texttt{Jaffe}      & 82.26 (0.07) & \underline{\textbf{82.25 (0.00)}} & 80.77 (0.90) & 78.72 (0.08) \\
\texttt{Isolet}     & 57.47 (0.03) & \underline{\textbf{60.09 (0.01)}} & 43.12 (1.21) & 58.61 (0.01) \\
\texttt{COIL20}     & 62.04 (0.03) & 61.02 (0.00)                   & 52.40 (1.59) & \underline{\textbf{63.82 (0.01)}} \\
\texttt{WarpPIE10P} & 28.25 (0.02) & \underline{\textbf{29.82 (0.03)}} & 28.03 (0.61) & 19.08 (0.03) \\
\texttt{USPS}       & 57.28 (0.02) & \underline{\textbf{65.78 (0.01)}} & 61.55 (1.17) & 65.24 (0.36) \\
\midrule
\multirow{2}{*}{Dataset} 
  & \multicolumn{4}{c}{Normalized Mutual Information (NMI) (\%)} \\
\cmidrule(lr){2-5}
  & Baseline & RAEUFS & URAFS & NNSE \\
\midrule
\texttt{lung}       & 56.64 (0.06) & \underline{\textbf{59.26 (0.02)}} & 55.40 (2.48) & 57.12 (0.27) \\
\texttt{Jaffe}      & 88.32 (0.03) & \underline{\textbf{88.01 (0.01)}} & 87.01 (0.61) & 79.32 (0.13) \\
\texttt{Isolet}     & 73.45 (0.02) & \underline{\textbf{75.34 (0.01)}} & 60.75 (1.04) & 72.61 (0.00) \\
\texttt{COIL20}     & 76.35 (0.01) & \underline{\textbf{75.88 (0.01)}} & 70.47 (1.33) & 75.52 (0.02) \\
\texttt{WarpPIE10P} & 29.80 (0.03) & \underline{\textbf{33.23 (0.01)}} & 29.24 (1.04) & 10.58 (0.01) \\
\texttt{USPS}       & 55.80 (0.02) & 61.12 (0.01)                   & 58.08 (0.84) & \underline{\textbf{61.50 (0.50)}} \\
\bottomrule
\end{tabular}%
}
% \vspace{5pt}

\end{table}

\begin{table}[!ht]
\centering

\caption{Comparison of RAEUFS, URAFS, and NNSE with 200 selected features for \textbf{dataset with 30\% outliers}. Mean and standard deviations (in parentheses) of ACC and NMI are reported over 100 repetitions. The best performance across the algorithms for each dataset is bold underlined.
}
\label{tab:outlier_result}

\resizebox{\linewidth}{!}{%
\begin{tabular}{lcccc}
\toprule
\multirow{2}{*}{Dataset} & \multicolumn{4}{c}{Accuracy (ACC) (\%)} \\
\cmidrule{2-5}
& Baseline & RAEUFS & URAFS & NNSE \\
\midrule
\texttt{lung}       & 69.46 (2.96)  & \underline{ \textbf{63.71 (2.84)}} & 62.98 (2.64) & 58.60 (0.61) \\ 
\texttt{jaffe}      & 34.74 (7.18) & \underline{ \textbf{89.76 (0.86)}} & 81.80 (1.00) & 72.57 (0.53) \\
\texttt{Isolet}     & 30.44 (5.28) & \underline{ \textbf{61.95 (1.09)}} & 50.62 (2.49) & 42.34 (0.13) \\
\texttt{COIL20}     & 42.22 (6.37) & \underline{ \textbf{66.85 (0.99)}} & 57.94 (0.85) & 60.67 (0.25) \\
\texttt{WarpPIE10P} & 14.05 (2.05) & 28.64 (0.96) & 28.03 (0.67) & \underline{\textbf{30.19 (0.19)}} \\
\texttt{USPS}         & 58.27 (0.05)  & \underline{\textbf{64.56 (0.00)}} & 60.81 (1.05)  & 63.59 (0.00) \\
\midrule
\multirow{2}{*}{Dataset} & \multicolumn{4}{c}{Normalized Mutual Information (NMI) (\%)} \\
\cmidrule{2-5}
& Baseline & RAEUFS & URAFS & NNSE \\
\midrule
\texttt{lung}       & 4.80 (14.39) & \underline{ \textbf{57.12 (1.72)}} & 53.84 (2.23) & 49.11 (0.61) \\
\texttt{jaffe}      & 55.69 (8.52) & \underline{ \textbf{91.28 (0.61)}} & 87.81 (0.61) & 76.21 (0.36) \\
\texttt{Isolet}     & 63.58 (5.22) & \underline{ \textbf{75.99 (0.85)}} & 67.56 (1.48) & 60.46 (0.09) \\
\texttt{COIL20}     & 65.35 (4.19) & \underline{ \textbf{78.20 (0.74)}} & 72.96 (0.51) & 73.60 (0.17) \\
\texttt{WarpPIE10P} & 4.33 (3.37)  & 30.59 (1.66) & 28.21 (1.09) & \underline{\textbf{31.49 (0.33)}} \\
\texttt{USPS}       & 57.33 (0.03)  & \underline{\textbf{60.45 (0.00)}} & 57.73 (0.94)  & 60.19 (0.00) \\
\bottomrule
\end{tabular}%
}

\end{table}

\begin{figure}[htbp]
    \centering
    \includegraphics[width=0.5\textwidth]{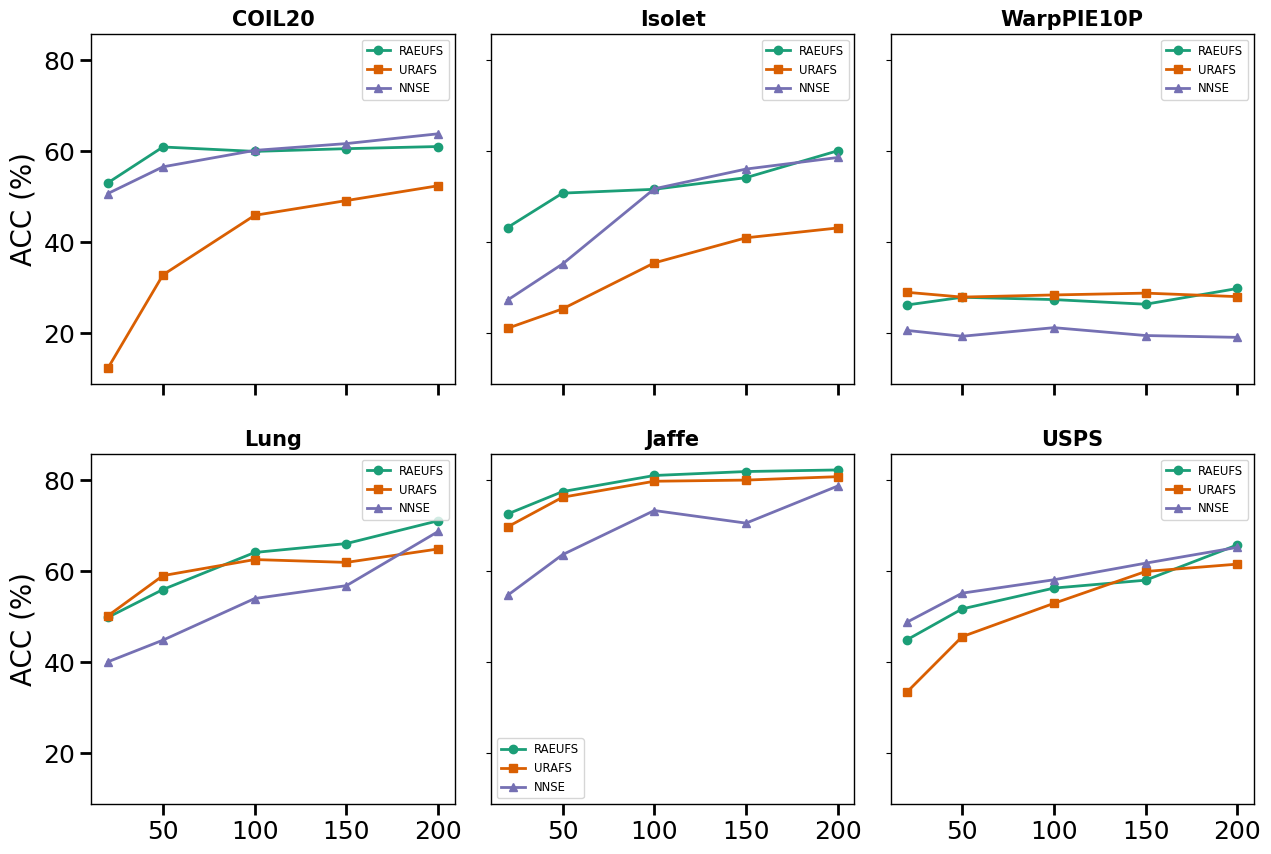}
    \caption{Performance of RAEUFS, URAFS and NNSE vs $\#$ selected features.}
    \label{fig:performance_comparison}
\end{figure}

Moreover, Fig. \ref{fig:performance_comparison} graphically illustrates the model’s stability and performance across the different number of selected features. On every dataset, RAEUFS is applied along with the other 2 methods with different number of selected features (20, 50, 100, 150, 200) and record their clustering ACC scores. RAEUFS consistently achieved higher or matched with the highest ACC scores, indicating that it identifies more relevant features that contribute meaningfully to clustering.

\begin{figure}[htbp]
    \centering
    \includegraphics[width=0.44\textwidth]{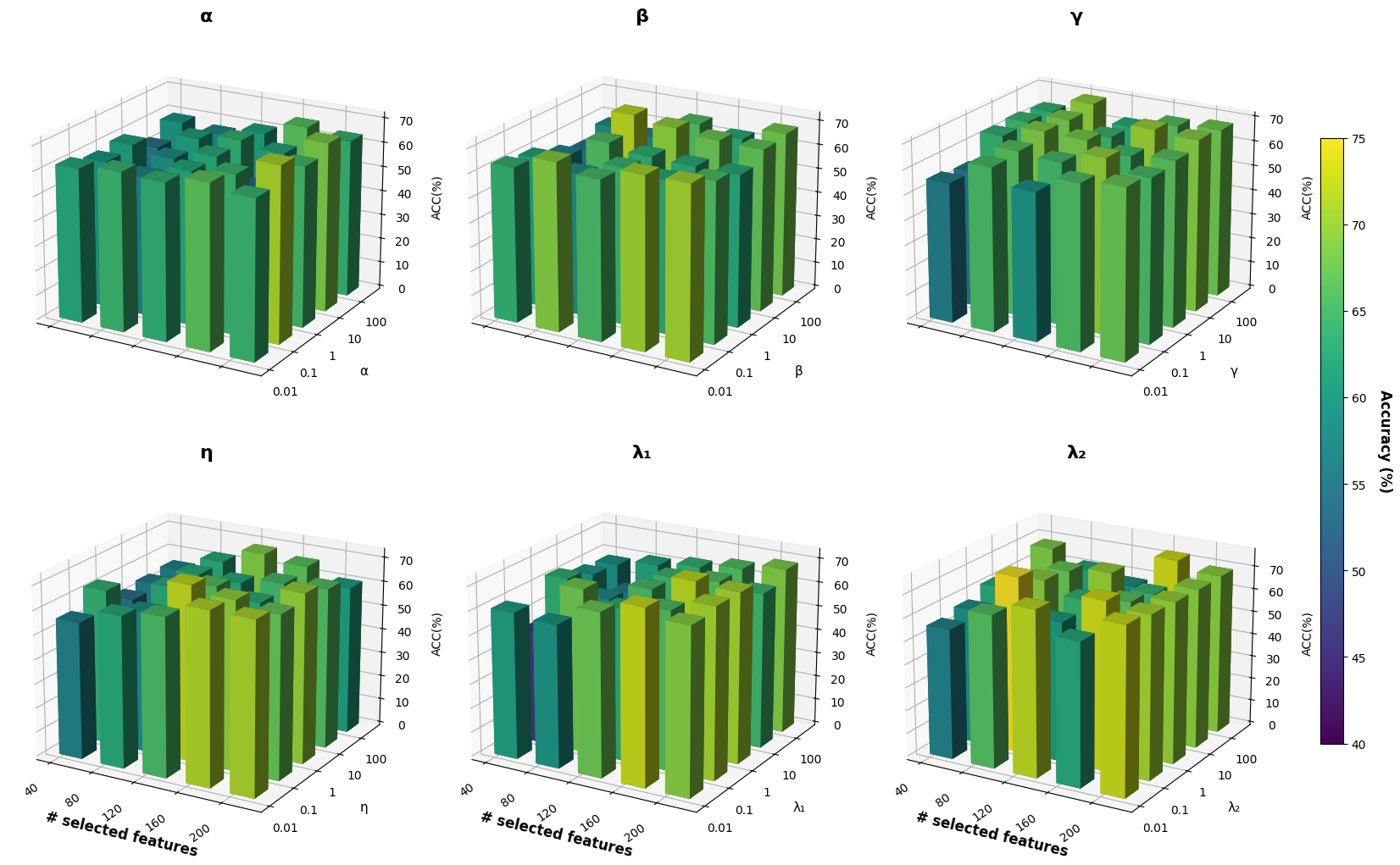}
    \caption{ACC sensitivity of $\alpha,\beta,\gamma,\eta,\lambda_1,\lambda_2$ on \texttt{lung}. 
    % {\color{red} the ticks were not fully shown, make the ticks font size larger, color legend}
    }
    \label{fig:Sensitivity_LUNG}
\end{figure}

% \subsection{Parameter sensitivity} \label{app:parameter_sensitivity}
\noindent\textbf{Parameter sensitivity.}
We additionally conduct a sensitivity analysis on \texttt{lung} to assess how sensitive RAEUFS is to its hyperparameters. A grid search of the parameters $(\alpha,\beta,\gamma,\eta,\lambda_1,\lambda_2) \times\#$ selected features is performed and the results of ACC are presented in Fig. \ref{fig:Sensitivity_LUNG}. Although the parameters vary logarithmically (base 10), the corresponding accuracy fluctuates irregularly. This suggests that determining optimal parameters for a new dataset remains challenging without comprehensive experimental validation.

\subsection{Dataset without Ground Truth: Migration Study across U.S.-Mexico border.}

To demonstrate the practical applicability of RAEUFS to complex real-world data, we evaluated the method on a dataset involving Mexican undocumented migration across the U.S.-Mexico border. The dataset stems from a survey on migration along Mexico's northern border (EMIF Norte)~\cite{EMIFNorte2019}, and consists of survey-based individual-level responses ($n=3,706$) capturing self-reported risks experienced during border crossings (e.g., abandonment, extreme temperatures, assault, etc). The primary objective was to capture latent structures that differentiate risk profiles across multiple cities, each with varying degrees of risk exposure. Notably, the dataset represents multiple challenges: heterogeneous feature types, measurement error, nonlinearity in feature interactions, and structural outliers. These characteristics make the dataset an ideal stress test for the robustness properties of RAEUFS.    
We hypothesize that Robust Autoencoder-based Unsupervised Feature Selection (RAEUFS) can reduce the dimensionality of the EMIF 
Norte dataset without compromising the clustering performance. 
Specifically, we hypothesize that by selecting the most informative 
features, our algorithm will achieve similar or improved clustering 
results compared to using the full set of features.

\begin{figure}[htbp]
    \centering
    \includegraphics[width=0.4\textwidth]{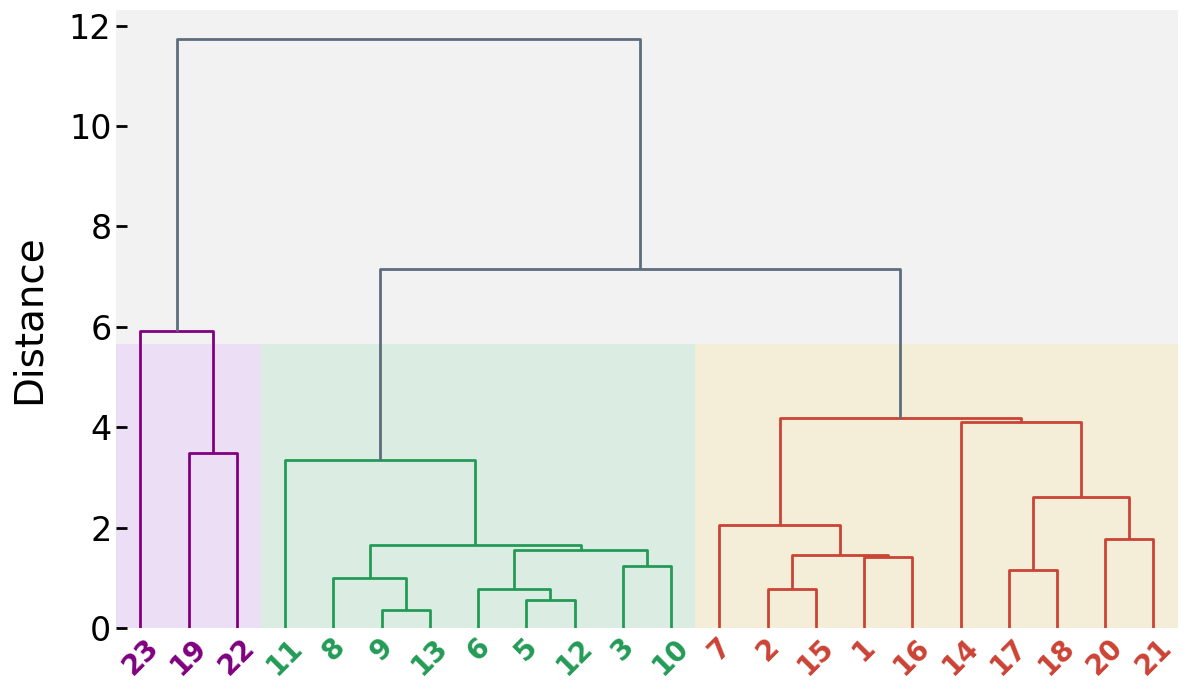}
    \caption{Dendrogram of the cities based on linkage distance.}
    \label{fig:real_life_dendogram}
\end{figure}

The unsupervised analysis was performed on nine risk factors, grouped by city of crossing. During preprocessing, we aggregated the dataset by city, computed the average risk values for each city, and applied robust-scaler normalization. Hierarchical clustering using all nine features is then applied (Fig. \ref{fig:real_life_dendogram}), which lead to three distinct clusters. However, \textit{k}-means found $2$ clusters as optimal based on Silhouette scores.
%while one cluster has only one city. 
% However, \textit{k}-means found $2$ clusters as optimal based on the Fig. \ref{fig:real_life_kmeans}. 

% \begin{figure}[htbp]
%     \centering
%     \includegraphics[width=0.35\textwidth]{fig/kmeans_silhoutte.png}
%     \caption{Silhouette score evaluation for \textit{k}-means clustering, indicating an optimal partition at \textit{k}=2.}
%     \label{fig:real_life_kmeans}
% \end{figure}

% \begin{table*}[ht]
% \centering
% \resizebox{\linewidth}{!}{%
% \begin{tabular}{cccccccc}
% \hline
% \textbf{cluster} & \textbf{h\_animals} & \textbf{h\_cold\_heat} & \textbf{h\_fall} & \textbf{h\_assaulted} & \textbf{h\_abandoned} \\ \hline
% 1 & 0.18 & 0.39 & 0.19 & 0.07 & 0.49 \\ 
% 2 & 0.06 & 0.40 & 0.12 & 0.03 & 0.17 \\ \hline
% \end{tabular}
% }
% \caption{K-means clustering results with 6 selected features by RAEUFS}
% \label{table:clustering}
% \end{table*}

\begin{table}[ht]
\centering

\caption{Silhouette scores for hierarchical clustering (HC) and \textit{k}-means with $k=2$ and $k=3$ by 
various numbers of selected features selected by RAEUFS.}
\label{tab:silhouette_scores}

% \resizebox{\linewidth}{!}{%
\begin{tabular}{ccccc}
\toprule 
& \multicolumn{2}{c}{\textbf{\textit{k}=2}} 
& \multicolumn{2}{c}{\textbf{\textit{k}=3}} \\
\cline{2-5} \\
\textbf{\# Features} & \textbf{HC} & \textbf{\textit{k}-means} 
& \textbf{HC} & \textbf{\textit{k}-means} \\
\midrule
3 & 0.46 & 0.44 & 0.34 & 0.29 \\
4 & \textbf{0.51} & 0.36 & 0.29 & 0.27\\
5 & \textbf{0.51} & \textbf{0.57} & 0.29 & 0.29 \\
6 & 0.46 & 0.46 & \textbf{0.41} & 0.26 \\
7 & 0.50 & 0.50 & 0.29 & 0.29 \\
8 & 0.44 & 0.44 &\textbf{ 0.41} & \textbf{0.38} \\
9 (all) & \textbf{0.51} & \textbf{0.57} & 0.29 & 0.32 \\
\bottomrule
\end{tabular}
% }

\end{table}

\Cref{tab:silhouette_scores} demonstrates the results of 
\textit{k}-means and Hierarchical clustering after 
selecting the features from our proposed method. Using all 9 features for \textit{k}=2, the hierarchical and \textit{k}-means clustering achieved silhouette scores of $0.51$ and $0.57$, respectively. This is our baseline score and we can see that by using only $4$ features (selected by the proposed RAEUFS algorithm) we achieved the same result as our baseline for the hierarchical clustering. However, the highest clustering performance was achieved with $5$ features, indicating that the algorithm successfully identified the most important features, leading to improved or comparable clustering results (\Cref{tab:silhouette_scores}). These clustering results reveal how unsupervised learning methods can reveal meaningful spatial variation on complex social phenomena, which supports data-informed intervention and sociologically grounded understanding of how risk emerges and how it can vary across geographical domains.

\section{Conclusion}\label{sec:conclusion}

We propose RAEUFS, a novel model for the UFS problem that integrates a Robust Subspace Recovery Autoencoder (RSR-AE) into the adaptive graph learning framework of embedding-based UFS methods. Leveraging the robustness of RSR-AE, RAEUFS effectively handles data contamination and consistently achieves the highest accuracy and NMI compared to existing methods. At the same time, it successfully identifies meaningful features across both synthetic and real-world datasets.

Several avenues for future research include: $\bullet$ Exploring adaptations of other robust autoencoder methods to the proposed framework. $\bullet$ Extending the framework to handle high-dimensional datasets with very small sample sizes could further broaden its applicability in fields such as genomics or finance. $\bullet$ Exploring online or incremental versions of RAEUFS would make the method suitable for streaming data scenarios, opening opportunities in real-time analytics and dynamic environments.

% \section{Acknowledgment}

% \textcolor{red}{If needed}

\bibliographystyle{IEEEtran}
\bibliography{ref-unsup-feature-selection}

\end{document}